\documentclass[11pt]{article}

\usepackage[preprint]{acl}

\usepackage{times}
\usepackage{latexsym}

\usepackage[T1]{fontenc}

\usepackage[utf8]{inputenc}

\usepackage{microtype}

\usepackage{inconsolata}

\usepackage{graphicx}

\usepackage{enumitem}

\usepackage{amsmath}
\usepackage{amsfonts}
\usepackage{booktabs}
\usepackage{multicol}
\usepackage{multirow}
\usepackage{tabularx}
\usepackage{dblfloatfix}

\usepackage{needspace}
\usepackage{enumitem}
\usepackage[ruled,vlined,linesnumbered]{algorithm2e}
\usepackage{setspace}
\usepackage{xspace}
\usepackage{xcolor}
\usepackage[most]{tcolorbox}
\definecolor{softred}{RGB}{250,100,100}
\definecolor{softgreen}{RGB}{56,118,29}
\definecolor{softblue}{RGB}{135,206,250}
\definecolor{softgray}{RGB}{150,150,150}

\newtcolorbox[auto counter, number within=section]{prompt}[2][]{%
  colback=white, 
  colframe=softblue!150, 
  width=\textwidth, 
  arc=3mm, 
  boxrule=0.8mm, 
  title=\normalsize #2, 
  fonttitle=\small, 
  fontupper=\footnotesize, 
  #1 
}

\usepackage{xspace}
\newcommand{\ourmemory}{\textsc{SDAM}\xspace}
\newcommand{\ourmethod}{\textsc{SDAM-SQL}\xspace}

\title{\ourmemory: Structure-Difference-Aware Memory Evolution for Complex Text-to-SQL}

\author{
Keyan Xu, Dingzirui Wang, Xuanliang Zhang, Qingfu Zhu, Wanxiang Che\thanks{Corresponding author.} \\
Harbin Institute of Technology \\
\{kyxu, dzrwang, xuanliangzhang, qfzhu, car\}@ir.hit.edu.cn
}

\begin{document}
\maketitle
\begin{abstract}
Text-to-SQL aims to convert natural language questions into executable SQL queries. While memory-based agent system improves complex SQL generation, existing memory design neglect historical experience and suffer from weak structure analysis, shallow semantic understanding, and poor schema alignment. To address these challenges, we propose \ourmemory. Specifically, \ourmemory identifies potential errors via a structure-difference aware reasoning tree, extracts deep semantic rules through contradiction-aware reflection, and enhances structural consistency using a schema-grounded memory evolution mechanism to bind memory with database schemas. We integrate \ourmemory into a Text-to-SQL framework named \ourmethod. 
Experiment shows that \ourmethod achieves $2.0\%$ and $0.4\%$ improvement on BIRD-dev and Spider-test compared with mainstream Text-to-SQL methods, showing the effectiveness of \ourmethod.

\end{abstract}

\section{Introduction}
    \begin{figure*}[t]
    \centering
    \includegraphics[width=1\textwidth]{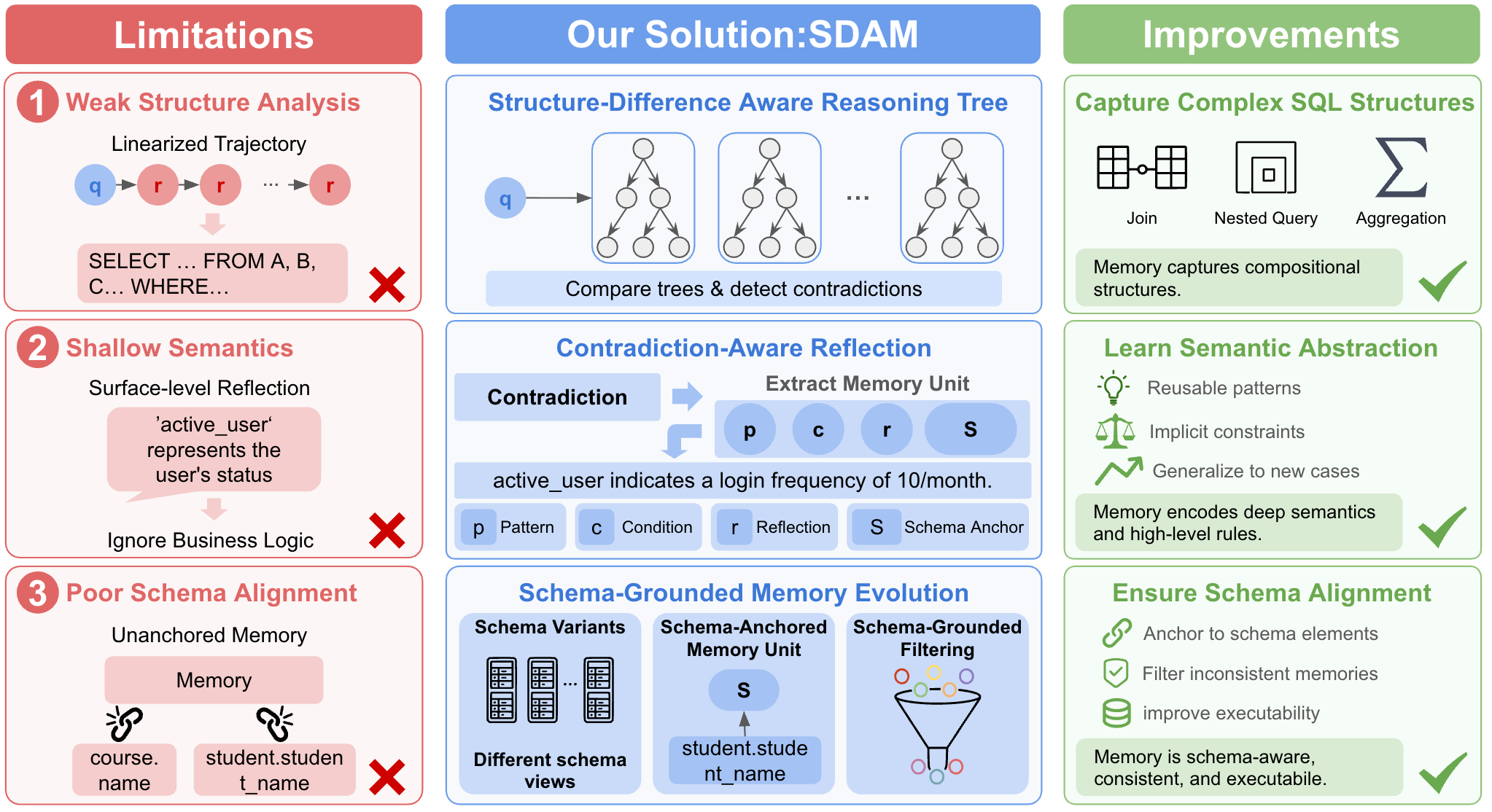}
    \caption{Limitations of existing methods and the motivation of \ourmemory. Existing methods suffer from three issues: \textit{(i) \textbf{Weak Structure Analysis}}, addressed by the Structure-Difference Aware Reasoning Tree (§\ref{sec:div_gen}); \textit{(ii) \textbf{Shallow Semantic Understanding}}, addressed by Contradiction-Aware Reflection (§\ref{sec:con_ref}); and \textit{(iii) \textbf{Poor Schema Alignment}}, addressed by the Schema-Grounded Memory Evolution Mechanism (§\ref{sec:str_ext}).}
    \label{fig:short}
\end{figure*}

Text-to-SQL aims to translate natural language questions into executable SQL queries \cite{baek2025knowledge}. 
Currently, agent-based frameworks have become the mainstream paradigm for complex SQL generation \cite{liu2025survey}. These methods improve Text-to-SQL performance through multi-stage collaborative reasoning. For example, CHESS \cite{talaei2024chess} employs multi-agent collaboration for SQL generation, while Alpha-SQL \cite{li2025alpha} leverages MCTS to search for optimal reasoning paths. However, most existing Agent-for-SQL methods still rely on static reasoning pipelines and lack continuous and effective utilization of historical experience, making adaptive optimization difficult in complex scenarios. To address this issue, memory mechanisms should be introduced to store historical experience. Recent general-purpose memory methods include the hierarchical graph memory of G-Memory \cite{zhang2025g} and the experience replay mechanism of ExpeL \cite{zhao2024expel}.

Although existing methods have achieved promising progress, their memory design still struggle to meet the demands of complex reasoning in Text-to-SQL, mainly in three aspects: \textit{(i) \textbf{Weak Structure Analysis}}: existing methods typically only record linear reasoning processes, making it difficult to model and analyze complex SQL logic such as multi-table joins and aggregation operations; \textit{(ii) \textbf{Shallow Semantic Understanding}}: existing methods mostly remain at surface-level semantic analysis and fail to capture the deeper semantics of database fields. For example, \texttt{active\_user} not only represents a user’s status, but also implicitly indicates that the user’s login frequency exceeds a certain threshold (Figure~\ref{fig:short}); \textit{(iii) \textbf{Poor Schema Alignment}}: existing memory lacks explicit alignment with the database schema, which can lead to SQL queries inconsistent with the database structure. For example, when the memory records “prefer using the \texttt{name} field when querying user names,” the model may incorrectly use \texttt{course.name} as the user name field, while the correct field should be \texttt{student.student\_name} (Figure~\ref{fig:short}).

To address these issues, we propose \textbf{S}tructure-\textbf{D}ifference-\textbf{A}ware \textbf{M}emory (\ourmemory), which achieves: 
\textit{(i) Weak Structure Analysis}: We introduce the \textit{Structure-Difference Aware Reasoning Tree}, which generates multiple reasoning paths and organizes them into a reasoning tree to capture structural differences among candidate SQL queries, thereby improving the accurate modeling of complex SQL structures. 
\textit{(ii) Shallow Semantic Understanding}: We propose \textit{Contradiction-Aware Reflection}, which analyzes logical contradictions in the reasoning process, inconsistencies between SQL and schema, as well as mismatches between execution results and question semantics, enabling the extraction of deep semantic rules from erroneous cases. 
\textit{(iii) Poor Schema Alignment}: We design the \textit{Schema-Grounded Memory Evolution Mechanism}, which explicitly binds memory to database tables and columns to reduce field confusion and structural errors.


Based on \ourmemory, we propose \ourmethod, a memory evolution framework for Text-to-SQL. 
\ourmethod leads to $0.4\%$ and $2.0\%$ improvement compared with existing mainstream Text-to-SQL methods, showing its effectiveness. 

Our contributions are summarized as follows:
\begin{itemize}[leftmargin=*,nosep]
    \item We propose \ourmemory, which achieves the evolution from instance-level errors to structured semantic knowledge through structure-difference analysis and contradiction-driven pattern extraction for complex Text-to-SQL reasoning tasks.
    \item Experiments on Spider and Bird benchmarks show that our method consistently outperforms strong baselines with $0.4\%$ and $2.0\%$, further demonstrating its effectiveness.
    \item Extensive analysis demonstrates that \ourmethod effectively alleviates the critical limitations of existing methods, achieving a compelling $4.04\%$ performance leap alongside improved computational efficiency compared to the state-of-the-art general-purpose memory framework, which firmly substantiates its domain-specific advantages in Text-to-SQL scenarios.
\end{itemize}

\section{Related Work}
    
\begin{figure*}[t]
    \centering
    \includegraphics[width=1\textwidth]{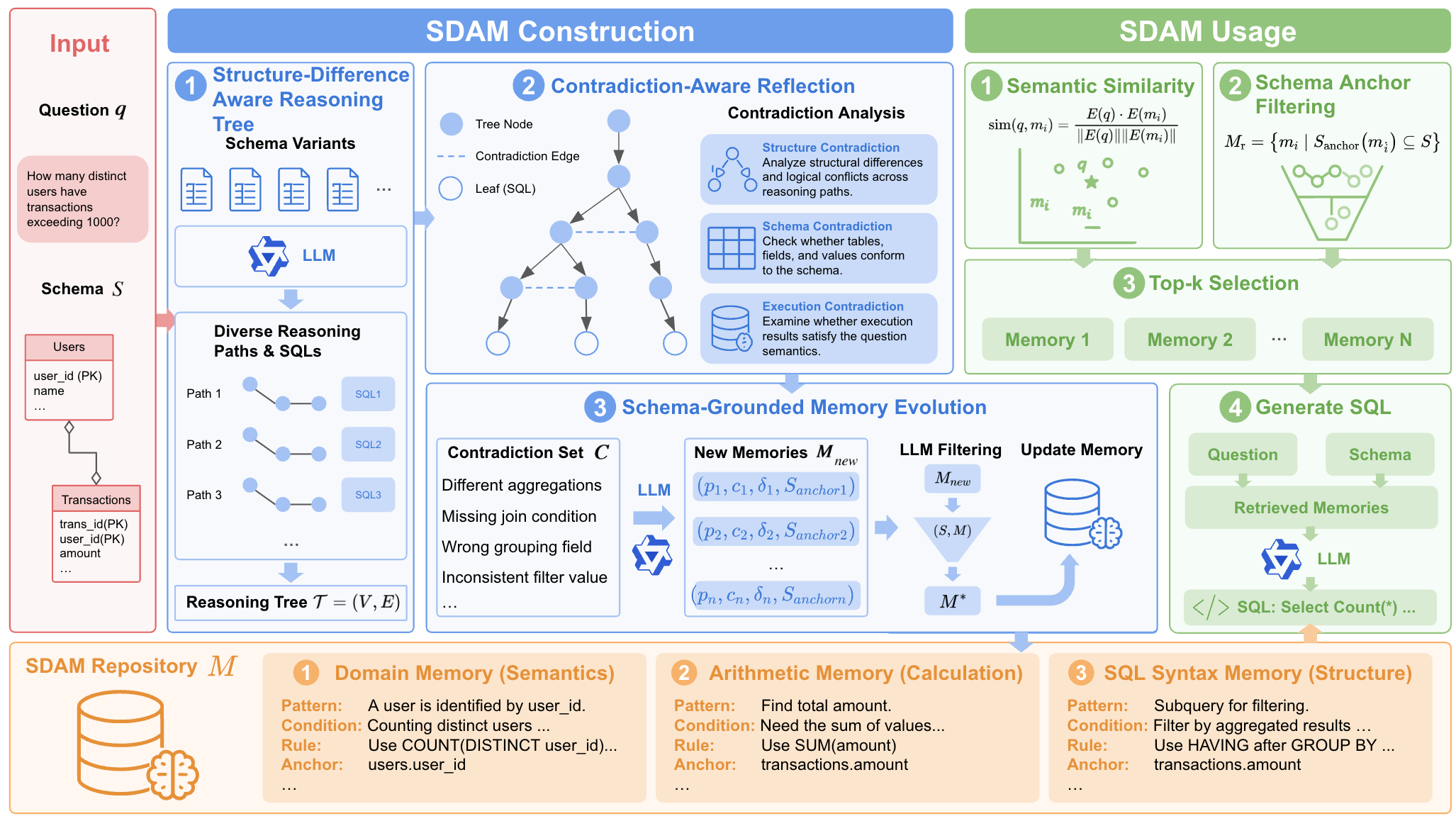}
\caption{\textbf{Overview of \ourmethod.} The workflow consists of two phases: 
    (a) \textbf{Memory Construction}: We generate diverse SQL candidates, detect contradictions via structure-difference aware reasoning tree, contradiction-aware reflection and schema-grounded memory evolution mechanism (\S\ref{sec:mem_con}). 
    (b) \textbf{Memory Usage}: Relevant memory units are dynamically retrieved via semantic similarity and schema anchors to guide SQL generation (\S\ref{sec:mem_usa}).}
    \label{fig:method}
\end{figure*}

\subsection{LLM-Powered Text-to-SQL}


Large Language Models (LLMs) have achieved significant progress in Text-to-SQL tasks due to their strong semantic understanding and code generation capabilities \citep{yang2025qwen3, achiam2023gpt, grattafiori2024llama, li2023can}. Recent studies mainly improve SQL generation through prompt engineering, task decomposition, and multi-path reasoning \citep{tai2023exploring, liu2023divide}.

The early methods relied on sophisticated prompting strategies, such as Chain-of-Thought (CoT) \cite{wei2022chain} and self-consistency reasoning \cite{wang2022self}. Later, DIN-SQL \cite{pourreza2023din} significantly improved complex SQL generation through task decomposition, while MCS-SQL \cite{lee2025mcs} enhanced robustness via multi-prompt search. To further improve complex reasoning ability, recent works introduced multi-agent collaboration and search-based mechanisms. For example, CHESS \cite{talaei2024chess} employs multi-agent collaboration for high-accuracy SQL generation, while Alpha-SQL \cite{li2025alpha} iteratively explores optimal reasoning paths through Monte Carlo Tree Search (MCTS), achieving performance on complex SQL tasks.

\subsection{Memory Evolution Agents}



Existing memory mechanisms for enhancing LLM agents in complex reasoning tasks can generally be categorized into three types: (1) parametric memory, which stores experience through fine-tuning or external parameter modules for continual adaptation \citep{chen2023fireact, yin2024agent, tack2024online}; (2) retrieval-based memory, which abstracts experience into reusable knowledge or skills for long-term adaptation \citep{zhang2025g, zhao2024expel, zheng2025skillweaver, wang2025mobile}; and (3) implicit memory, which encodes and retrieves historical experience through latent representations \citep{wang2025m+, hu2025beyond}.

Among them, retrieval-based memory has recently attracted increasing attention due to its interpretability and scalability. 
G-Memory \cite{zhang2025g} organizes multi-agent experiences with hierarchical graph structures to support collaborative reasoning; 
ExpeL \cite{zhao2024expel} stores historical trajectories and natural language rules through experience replay for iterative self-improvement; 
SkillWeaver \cite{zheng2025skillweaver} abstracts procedural experience into reusable APIs and executable skills; 
and Mobile-Agent-E \cite{wang2025mobile} continuously improves mobile agents through long-term memory mechanisms. 

\section{Task Formulation}


The Text-to-SQL task aims to map a natural language question into an executable SQL query. Formally, given a question $q$ and a database schema $S = \{T_1, \ldots, T_m\}$ where each table contains a set of columns $T_i = \{c_{i1}, \ldots, c_{in}\}$, the goal is to generate the SQL query via $y = f(q, S)$. 

Distinguishing from conventional methods, we introduce and maintain a dynamically evolving, structured memory set $M = {m_i}$ to store structural patterns, semantic rules, and schema-aligned knowledge extracted from complex reasoning processes. Consequently, we reformulate the SQL generation task as: $y = f(q, S, M)$ where $M$ injects supplementary structural expertise and semantic constraints into the generation process, thereby enhancing both the execution correctness and schema consistency of the synthesized queries.
\section{Methodology}


This section presents \ourmemory, a structured memory framework for Text-to-SQL that conceptualizes memory as the artifact of structural discrepancy and contradiction analysis. The pipeline consists of two primary phases (Figure~\ref{fig:method}): (1) \textbf{\ourmemory Construction}, which builds dynamically evolving, structured memories via structural difference analysis and contradiction-driven pattern extraction; and (2) \textbf{\ourmemory Usage}, which retrieves and enforces memory constraints via semantic similarity and schema anchors to guide SQL generation. 
\ourmemory systematically resolves three core challenges: (1) \textbf{Weak Structure Analysis} via structure-difference aware reasoning tree (\S\ref{sec:div_gen}) to model complex SQL structures; (2) \textbf{Shallow Semantic Understanding} via contradiction-aware reflection (\S\ref{sec:con_ref}) to capture latent column semantics; and (3) \textbf{Poor Schema Alignment} via schema-grounded memory evolution mechanism (\S\ref{sec:str_ext}) to ensure explicit memory-to-schema correspondence.

\subsection{\ourmemory Construction}
\label{sec:mem_con}
\subsubsection{Structure-Difference Aware Reasoning Tree}
\label{sec:div_gen}


To address the weak structure analysis problem, \ourmemory constructs a structure-difference aware reasoning tree. Specifically, we first leverage the high sensitivity of LLMs to structural variations in schema representations \cite{liu2026xiyan} by constructing multiple schema variants through modifying field descriptions and schema organization forms (See Appendix \ref{app:schema} for details):
\[
\tilde{S} = \{S_k\}_{k=1}^{m}
\]
Based on different schema variants, the model generates corresponding SQL candidates $y_k$ and reasoning paths $R_k$ in parallel under the contextualized guidance of structured memory $M$:
\[
(y_k, R_k) = f(q, S_k, M), \quad k = 1,\ldots,m
\]

Since SQL reasoning naturally exhibits staged and structured characteristics \cite{li2025alpha}, we regard each reasoning path $R_k$ as a sequence of structured reasoning nodes organized in a fixed reasoning order (see Table~\ref{tab:Nodes}), including question rephrasing, table selection, column selection, function identification, condition value identification, and SQL generation.

After obtaining multiple reasoning paths, we further topologically merge semantically identical nodes across different paths while preserving structurally different or logically conflicting parts as divergent, independent branches, thereby constructing a unified reasoning tree:
\[
\mathcal{T} = (V, E)
\]
where $V$ denotes the set of reasoning nodes and $E$ denotes the dependency relations between nodes. In this way, \ourmemory can explicitly model structural differences among candidate SQL queries, enabling more effective detection of potential structural errors in complex SQL reasoning.

\subsubsection{Contradiction-Aware Reflection}
\label{sec:con_ref}



To address the shallow semantic understanding problem, \ourmemory further introduces contradiction-aware reflection based on the reasoning tree $\mathcal{T}$. Different from conventional surface-level reflection that relies only on a single SQL query, \ourmemory leverages LLMs to jointly analyze the intersection nodes and conflicting branches within the reasoning tree. Under the organization of the reasoning tree, structural differences and logical conflicts across different reasoning paths are explicitly exposed, making potential contradictions clearer and easier for the model to identify and analyze, thereby uncovering potential errors and implicit semantic information in complex SQL reasoning.

Specifically, contradictions are analyzed from three complementary perspectives at different granularities (See Appendix \ref{app:con_det} for more details):
\begin{itemize}[leftmargin=*,nosep]
    \item \textbf{Structure Contradiction}: captures structural differences and logical conflicts among different reasoning paths;
    \item \textbf{Schema Contradiction}: checks the consistency between generated SQL and schema definitions;
    \item \textbf{Execution Contradiction}: determines whether the execution results satisfy the question semantics and constraints.
\end{itemize}

Finally, all detected contradictions are summarized into a global contradiction set:
\[
\mathcal{C} = LLM_{Reflector}\big(\mathcal{T}, \tilde{S}, q\big)
\]

By comparing fine-grained intermediate reasoning branches within the reasoning tree, \ourmemory can more effectively identify potential errors in complex divergent SQL reasoning.

\subsubsection{Schema-Grounded Memory Evolution}
\label{sec:str_ext}




To address the poor schema alignment issue, \ourmemory introduces a schema-grounded memory evolution mechanism, which consists of three stages: memory extraction, memory filtering, and memory evolution.

First, the system performs structured memory extraction based on the contradiction set $\mathcal{C}$, transforming the error patterns and semantic rules identified during reflection into a new memory set:
\[
M_{\text{new}} = LLM_{Extractor}(\mathcal{C})
\]
To achieve explicit schema alignment, each memory unit is defined as a quadruple:
\[
m = (p, c, \delta, S_{\text{anchor}})
\]
where $p$ denotes the structural or semantic pattern, $c$ denotes the triggering condition, $\delta$ denotes the correction rule, and $S_{\text{anchor}}$ represents the Schema Anchor, which explicitly binds the memory to tables and columns in the database, thereby reducing column confusion and incorrect schema mappings.

Next, considering that LLM-generated memories may contain noise or hallucinations, we further conduct consistency verification and filtering on the newly extracted memories by jointly leveraging the schema variant set $\tilde{S}$ and the existing memory set $M$, resulting in a high-confidence memory set:
\[
M^{*} = LLM_{Filter}(M_{\text{new}}, \tilde{S}, M)
\]

Finally, the memory bank is continuously updated in an incremental manner:
\[
M \leftarrow M \cup M^{*}
\]

To better handle different types of Text-to-SQL challenges, the retained high-quality memories are further categorized into three types:
\begin{itemize}[leftmargin=*,nosep]
    \item \textbf{Domain Memory}: stores domain semantics and business rules;
    \item \textbf{Arithmetic Memory}: models aggregation, ranking, and numerical computation patterns;
    \item \textbf{SQL Syntax Memory}: captures SQL structural patterns and multi-table join relationships.
\end{itemize}

\subsection{Memory Usage}
\label{sec:mem_usa}










During SQL generation, we perform dynamic retrieval over structured memory based on the semantic meaning of the query and the database schema, providing auxiliary information for reasoning.

First, we compute the semantic similarity between the query and each memory unit:
\[
\text{sim}(q, m_i) =
\frac{E(q)\cdot E(m_i)}
{\|E(q)\|\|E(m_i)\|}
\]
where $E(\cdot)$ denotes an embedding encoder that maps inputs into a shared semantic space.

Based on this, we introduce a schema-anchor constraint to filter out memory units that are not aligned with the current schema:
\[
M_{\text{align}} =
\{ m_i \mid S_{\text{anchor}}(m_i) \subseteq S \}
\]
where $S$ denotes the current database schema.

Within the aligned memory set, we select the top-$k$ most relevant memory units according to the similarity score:
\[
M_k = \text{TopK}(\text{sim}(q, m_i), M_{\text{align}})
\]
Finally, the retrieved memory is explicitly injected into the SQL generation process as context-aware, structured reasoning guidance, enhancing the model’s ability to capture complex semantic patterns and schema constraints.

\section{Experiment}
    \begin{table*}[t]
\centering
\small
\setlength{\tabcolsep}{6pt}
\renewcommand{\arraystretch}{1.2}
\begin{tabular}{l l c c}
\toprule
\textbf{Methods} & \textbf{Model} & \textbf{BIRD-EX (\%)} & \textbf{Spider-EX (\%)} \\
\midrule

\multicolumn{4}{c}{\textbf{Close-Source Model-Based Baselines}} \\
\midrule
DIN-SQL \cite{pourreza2023din} & GPT-4 & $50.7$ & $85.3$ \\
DAIL-SQL \cite{gao2023text} & GPT-4 & $55.9$ & $86.6$ \\
SuperSQL \cite{li2024dawn} & GPT-4 & $58.5$ & -- \\
RSL-SQL \cite{cao2024rsl} & GPT-4o & $67.2$ & $87.9$ \\
CHESS (IR,CG,UT) \cite{talaei2024chess} & Gemini-1.5-Pro & $68.3$ & -- \\

\midrule
\multicolumn{4}{c}{\textbf{Open-Source Model-Based Baselines}} \\
\midrule
GBV-SQL \cite{chen2025gbv} & Deepseek-v3 & $63.2$ & $79.6$ \\
RSL-SQL \cite{cao2024rsl} & Deepseek-v2 & $63.6$ & $87.5$ \\
GenaSQL \cite{donder2025cheaper} & Qwen3-Coder-30B-A3B-Instruct & $66.9$ & $87.6$ \\
ExpeSQL \cite{zeruiexpesql} & Qwen3-Coder-30B-A3B-Instruct & $67.5$ & -- \\
Alpha-SQL \cite{li2025alpha} & Qwen3-Coder-30B-A3B-Instruct & $68.2$ & -- \\

\midrule
\multicolumn{4}{c}{\textbf{Ours}} \\
\midrule
\textbf{\ourmethod} & \textbf{Qwen3-Coder-30B-A3B-Instruct} & $\mathbf{70.2}$ & $\mathbf{88.0}$ \\

\bottomrule
\end{tabular}
\caption{Execution accuracy comparison on BIRD and Spider benchmarks. \textbf{Bold} indicates the best performance.}
\label{tab:overall_performance}
\end{table*}

\subsection{Experimental Setup}
\paragraph{Datasets}
We evaluate our proposed method on two representative and challenging Text-to-SQL benchmarks: 
(1) \textbf{Spider} \cite{yu2018spider}, a classic cross-domain benchmark containing 10,181 questions and 5,693 unique SQL queries across 200 databases, highlighting the model's compositional generalization ability on unseen schemas; 
(2) \textbf{BIRD} \cite{li2023can}, a large-scale real-world benchmark with 12,751 question-SQL pairs across 95 large-sized databases, characterized by noisy data and intricate schemas that pose severe challenges for schema alignment and robustness.

\paragraph{Evaluation Metrics}
Following prior work \citep{cao2024rsl, liu2026xiyan, pourreza2024chase}, we adopt Execution Accuracy (EX) as our metric. EX measures correctness by executing both the predicted and gold SQL queries on the target database and verifying whether their execution results are semantically equivalent (ignoring row order). 

\paragraph{Baselines}




We compare \ourmethod with a suite of state-of-the-art, prompting-based Text-to-SQL methods. These approaches follow an in-context learning paradigm without task-specific fine-tuning, leveraging Large Language Models (LLMs) directly for SQL generation. According to the underlying backbone models, they are categorized into two methodological groups:

\textbf{Close-Source Model-Based Methods}: This category relies on advanced commercial models (e.g., GPT-4o, Gemini) and includes DIN-SQL \cite{pourreza2023din}, DAIL-SQL \cite{gao2023text}, SuperSQL \cite{li2024dawn}, RSL-SQL \cite{cao2024rsl}, and CHESS \cite{talaei2024chess}.

\textbf{Open-Source Model-Based Methods}: Built upon publicly available, highly scalable open-weights models (e.g., DeepSeek, Qwen-Coder), this group encompasses GBV-SQL \cite{chen2025gbv}, RSL-SQL \cite{cao2024rsl}, GenaSQL \cite{donder2025cheaper}, Alpha-SQL \cite{li2025alpha} and ExpeSQL \cite{zeruiexpesql}.

Additionally, we employ a vanilla SQL-generation framework (See Appendix \ref{app:gen_str} for details) as our baseline, upon which \ourmethod is built by integrating \ourmemory.

\paragraph{Implementation Details}
All experiments are conducted on an Ubuntu 22.04 server. We locally deploy all open-weights models via the vLLM framework \cite{kwon2023efficient} on two NVIDIA A100 (80GB) GPUs. To evaluate the generalizability across different foundation capabilities, we implement \ourmethod on various backbone models, including the Qwen3 series (8B/14B/32B), the Qwen2.5-Coder series (7B/14B/32B), and Qwen3-Coder-30B-A3B-Instruct \cite{yang2025qwen3,hui2024qwen2}, which drive \ourmemory construction and SQL generate (Appendix \ref{app:gen_str}) phases. 
We adopt the online adaptation setting, where methods are evaluated sequentially on the dev/test split. For each sample, the model first generates predictions based on the current context and then updates its memory according to the observed reasoning process. To ensure a fair comparison, \ourmemory is constructed and evolved autonomously without access to any ground-truth labels throughout the entire evaluation process.
For semantic value retrieval, we employ Qwen3-Embedding-0.6B \cite{zhang2025qwen3}. To enhance the diversity of reasoning paths, we construct five distinct schema representations for each database, generating a candidate SQL query from each representation to enrich the input for subsequent contradiction analysis and memory extraction. Prompt templates and detailed configurations are provided in the Appendix \ref{app:prompts}.

\begin{table*}[t]
\centering
\small
\setlength{\tabcolsep}{5pt}
\renewcommand{\arraystretch}{1.2}

\begin{tabular}{l l c c c c c c c}
\toprule

\multirow{2}{*}{\textbf{Dataset}} &
\multirow{2}{*}{\textbf{Method}} &
\multicolumn{3}{c}{\textbf{Qwen3}} &
\multicolumn{3}{c}{\textbf{Qwen2.5-Coder}} &
\textbf{Qwen3-Coder} \\

\cmidrule(lr){3-5}
\cmidrule(lr){6-8}
\cmidrule(lr){9-9}

&
&
\textbf{8B} &
\textbf{14B} &
\textbf{32B} &
\textbf{7B} &
\textbf{14B} &
\textbf{32B} &
\textbf{30B-A3B} \\

\midrule

\multirow{2}{*}{Spider-Test}
& Baseline
& $80.46$ & $81.14$ & $84.82$
& $83.17$ & $86.46$ & $\mathbf{85.59}$
& $87.62$ \\

& \ourmethod
& $\mathbf{81.00}$ & $\mathbf{81.72}$ & $\mathbf{85.69}$
& $\mathbf{84.23}$ & $\mathbf{87.12}$ & $85.50$
& $\mathbf{88.01}$ \\

\midrule

\multirow{2}{*}{BIRD-Dev}
& Baseline
& $57.04$ & $61.67$ & $64.02$
& $60.43$ & $64.71$ & $64.89$
& $66.88$ \\

& \ourmethod
& $\mathbf{60.04}$ & $\mathbf{63.30}$ & $\mathbf{65.19}$
& $\mathbf{62.71}$ & $\mathbf{66.73}$ & $\mathbf{66.88}$
& $\mathbf{70.21}$ \\

\bottomrule
\end{tabular}

\caption{Performance comparison between baseline and \ourmethod with memory evolution across different model scales and datasets. \textbf{Bold} indicates the best performance.}
\label{tab:main_results}

\end{table*}
\subsection{Overall Performance}

Table \ref{tab:overall_performance} presents the overall experimental results of \ourmethod on the BIRD and Spider benchmarks. The empirical results demonstrate that our method consistently surpasses existing baselines across all configurations on both datasets.

Specifically, on the highly challenging BIRD benchmark, \ourmethod configured with Qwen3-Coder-30B-A3B-Instruct yields an Execution Accuracy (EX) of \textbf{70.2\%}, achieving absolute improvements of \textbf{2.0}\% and \textbf{2.7}\% over the strongest baselines, Alpha-SQL ($68.2\%$) and ExpeSQL ($67.5\%$), respectively. On the Spider dataset, \ourmethod also secures a top-tier EX of \textbf{88.0\%}, outperforming competitive baselines such as GenaSQL ($87.6\%$) and RSL-SQL. Notably, the performance gain of \ourmethod is more pronounced on BIRD, which features denser complex scenarios, underscores its distinct advantage in overcoming the bottlenecks of complex SQL memory evolution and schema alignment through structural discrepancy analysis and the contradiction-driven memory mechanism.

\subsection{Main Results}

Table \ref{tab:main_results} presents the experimental results of \ourmethod across varying model scales and datasets. The empirical findings indicate that our method consistently robustly boosts Text-to-SQL performance under most model configurations, with the performance gains becoming more pronounced as the inherent task complexity scales up.

Specifically: (1) On the relatively well-regulated Spider dataset, most models exhibit stable improvements, where Qwen3-32B and Qwen2.5-Coder-14B increase to $85.69\%$ and $87.12\%,$ respectively, with the peak performance reaching \textbf{88.01\%}. (2) On the more challenging BIRD benchmark, all models achieve consistent gains; for instance, Qwen3-8B improves by \textbf{3.00}\%, and Qwen3-Coder-30B-A3B-Instruct reaches \textbf{70.21\%}, demonstrating \ourmethod's capability to navigate complex structures and implicit semantic constraints. (3) We observe similar consistent gains on the complex-reasoning Archer dataset (Appendix \ref{app:exp_arc}), where the performance peaks at \textbf{36.54\%}.These results validate the effectiveness of our memory evolution mechanism in complex Text-to-SQL tasks.
\subsection{Ablation Study}
To verify the effectiveness of each module, we conduct ablation studies on the BIRD benchmark, thereby isolating the unique contribution of each architectural component (Table \ref{tab:ablation}):
\begin{itemize}[leftmargin=*,nosep]
    \item \textbf{w/o tree}: Performance drops by \textbf{3.26}\%, confirming the critical role of the reasoning tree in capturing latent structural errors.
    \item \textbf{w/o contradiction}: The performance drops by \textbf{2.35}\%, indicating that the analysis of multi-dimensional contradictions effectively breaks the bottleneck of shallow semantic understanding.
    \item \textbf{w/o schema anchor}: Performance declines by \textbf{1.76}\%, demonstrating that explicitly binding memories to database elements effectively reduces schema mapping errors.
    \item \textbf{w/o filter}: Performance also degrades, validating the necessity of the filtering mechanism in suppressing noise and enhancing memory reliability.
\end{itemize}
In summary, all modules synergistically contribute positively to the overall performance, with structural discrepancy analysis and schema-anchored retrieval playing a particularly prominent role in supporting complex reasoning.

\begin{table}[t]
\small
\centering
\setlength{\tabcolsep}{8pt}
\renewcommand{\arraystretch}{1.2}

\begin{tabular}{l c c}
\toprule
\textbf{Configuration} & \textbf{EX (\%)} & $\mathbf{\Delta}$ \\
\midrule

\ourmethod & $\mathbf{70.21}$ & -- \\

\midrule

w/o tree          & $66.95$ & $-3.26$ \\
w/o contradiction & $67.86$ & $-2.35$ \\
w/o schema anchor & $68.45$ & $-1.76$ \\
w/o filter        & $69.75$ & $-0.46$ \\

\bottomrule
\end{tabular}

\caption{
Ablation study of different components in \ourmethod on the BIRD dataset.
}
\label{tab:ablation}

\end{table}
\begin{table}[t]
\small
\centering
\renewcommand{\arraystretch}{1.2}

\begin{tabular}{lccc}
\toprule
\textbf{Difficulty} & \textbf{Baseline} & \textbf{\ourmethod} & $\mathbf{\Delta}$ \\
\midrule

Simple      & $72.32$ & $\mathbf{74.59}$ & $\mathbf{+2.27}$ \\
Moderate    & $60.34$ & $\mathbf{64.01}$ & $\mathbf{+3.67}$ \\
Challenging & $53.10$ & $\mathbf{62.07}$ & $\mathbf{+8.97}$ \\

\bottomrule
\end{tabular}

\caption{Performance comparison across different SQL difficulty levels on BIRD. $\Delta$ denotes the improvement over the baseline.}
\label{tab:difficulty_analysis}

\end{table}
\begin{table}[t]
\centering
\small
\begin{tabular}{llcc}
\toprule
\textbf{Base Model} & \textbf{Method} & \textbf{EX (\%)} & $\mathbf{\Delta}$ \\
\midrule
\multirow{3}{*}{Qwen3-Coder} & baseline & $66.88$ & $0.00$ \\
 & ACE & $66.17$ & $-0.71$ \\
 & \textbf{\ourmethod} & $\mathbf{70.21}$ & $\mathbf{+3.33}$ \\
\midrule
\multirow{3}{*}{Qwen3-32B} & baseline & $64.02$ & $0.00$ \\
 & ACE  & $62.65$ & $-1.37$ \\
 & \textbf{\ourmethod} & $\mathbf{65.19}$ & $\mathbf{+1.17}$ \\
\bottomrule
\end{tabular}
\caption{Performance comparison between ACE \cite{zhang2025agentic} and \ourmethod. \textbf{Bold} indicates the best results; $\Delta$ represents the improvement over baseline.}
\label{tab:compare}
\end{table}
\begin{table}[t]
\centering
\small
\renewcommand{\arraystretch}{1.2}

\begin{tabular}{l c c c}
\toprule
\textbf{Method} & \textbf{LLM Calls} & \textbf{Tokens (K)} & \textbf{Time (s)} \\
\midrule

ACE & $10.5$ & $30.2$ & $4.7$ \\
\textbf{\ourmethod} & $\mathbf{8.6}$ & $\mathbf{26.6}$ & $\mathbf{4.0}$ \\

\bottomrule
\end{tabular}

\caption{
Efficiency comparison in terms of average LLM calls, token consumption, and execution time per query. \textbf{Bold} indicates the most efficient performance with lower computational overhead.
}
\label{tab:efficiency}

\end{table}
\subsection{Difficulty-Level Analysis}

Table \ref{tab:difficulty_analysis} presents the performance of \ourmethod across different SQL difficulty levels on the BIRD benchmark. The results indicate that the advantage of our method becomes increasingly pronounced as the task difficulty scaling up, yielding performance gains of \textbf{2.27}\%, \textbf{3.67}\%, and \textbf{8.97}\% under Simple, Moderate, and Challenging settings, respectively.

Challenging samples typically involve intricate structures such as multi-table joins and nested aggregations, where conventional generation strategies are highly prone to failure. In contrast, leveraging structural discrepancy analysis and the reasoning tree mechanism, \ourmethod precisely pinpoints latent structural defects and distills deep semantic rules through contradiction-driven memory evolution. This capability empowers the model with superior error correction and structural modeling in Challenging scenarios, thereby contributing to the most breakthrough performance leaps.

\subsection{Comparative Analysis}


Table \ref{tab:compare} presents the comparative results between the SOTA general memory framework ACE \cite{zhang2025agentic} and \ourmethod across various LLM backbones.

Experimental results demonstrate that after introducing the state-of-the-art general memory framework ACE, the performance of Qwen3-Coder-30B-A3B and Qwen3-32B decreases by $0.71\%$ and $1.37\%$, respectively. This indicates that although existing general-purpose methods can enhance the long-term experience utilization of agents, they lack the capability to model complex SQL structures, deep column semantics, and schema alignment, thereby making it difficult to effectively adapt to Text-to-SQL scenarios.

In contrast, leveraging structure-difference analysis, contradiction-driven memory extraction, and the Schema-Grounded evolution mechanism, \ourmethod effectively alleviates structural errors, semantic deviations, and column confusion in complex SQL reasoning. Specifically, \ourmethod brings a $3.33\%$ performance improvement to Qwen3-Coder (reaching $70.21\%$ EX) and also achieves a stable growth of $1.17\%$ on Qwen3-32B. This firmly substantiates that the structured memory evolution mechanism tailored specifically for Text-to-SQL tasks can more efficiently and significantly enhance the generation quality and execution correctness of complex SQL queries.

\subsection{Efficiency Analysis}
To provide a clear perspective on the computational cost, we evaluate the memory construction efficiency of \ourmethod against the general-purpose memory framework ACE on Qwen3-Coder-30B-A3B in Table~\ref{tab:efficiency}. Experimental results demonstrate that \ourmethod exhibits distinct advantages across all efficiency metrics. Specifically, in terms of the average LLM calls per query, \ourmethod substantially reduces the frequency from $10.5$ to $8.6$. This reduction is largely attributed to our carefully designed contradiction-driven reflection mechanism, which effectively eliminates blind and redundant iterations. Consequently, in terms of token consumption, \ourmethod saves approximately $11.9\%$ of overhead compared to ACE (dropping from $30.2\text{K}$ to $26.6\text{K}$). Profiting from the streamlined calling pipeline, the average execution time of \ourmethod is only $4.0$ seconds, achieving a nearly $14.9\%$ speedup over ACE ($4.7$ seconds). These findings firmly substantiate that while generating high-quality memory, \ourmethod achieves superior computational efficiency and significantly lower operational overhead.

\section{Conclusion}

In this paper, we propose \ourmemory, a structure-difference-aware memory method for Text-to-SQL, and build its evolution framework, \ourmethod. Specifically, \ourmemory leverages a Structure-Difference Aware Reasoning Tree, incorporates Contradiction-Aware Reflection, and introduces a Schema-Grounded Memory Evolution Mechanism to address key challenges in complex SQL memory evolution. Extensive experiments on the Spider, BIRD, and Archer datasets demonstrate the effectiveness of our approach in complex structural modeling, deep semantic rule extraction, and precise schema alignment. 
Moreover, compared with general-purpose memory frameworks, \ourmemory demonstrates clear advantages in both performance and computational efficiency, highlighting its importance for Text-to-SQL tasks.
In the future, we will explore memory generalization in cross-database transfer scenarios and integrate long-cycle feedback mechanisms to further enhance the system's self-evolution capabilities.

\section*{Limitation}
Although \ourmemory significantly improves performance across multiple Text-to-SQL benchmarks, it exhibits two primary limitations. First, the current memory evolution heavily relies on logical contradictions and error feedback during the reasoning process. Consequently, when encountering highly custom domain-specific knowledge or extremely rare business logic (i.e., cold-start scenarios), \ourmemory may struggle to unearth deep semantic rules purely through self-reflection without the guidance of external domain ontologies or expert priors. Second, because the Schema-Grounded Memory Evolution Mechanism explicitly binds memory units to specific database schemas, the generalization and transferability of these localized, structured memories to completely unseen databases with distinct structures remains an open challenge that requires further exploration.

\section*{Ethics Statement}
This work complies with standard ethical guidelines. All datasets utilized in our evaluation—Spider, BIRD, and Archer—are publicly accessible academic benchmarks that contain no private or personally identifiable information. The proposed framework is intended solely for research purposes to advance natural language interfaces for databases. To prevent risks such as accidental data modification or malicious SQL injection in practical applications, real-world deployment of automated SQL generation should always be equipped with strict privilege control and read-only execution sandboxes. We employ the LLM tool to polish the paper writing.

\bibliography{cite}

\clearpage
\appendix

\section{Generate Strategy}
\label{app:gen_str}




Our SQL generation strategy adopts the design from \cite{donder2025cheaper}, which conceptualizes the pipeline into four primary stages:

(1) Schema Linking Stage:
This stage utilizes multiple text formats to project the target database structure into diverse representational modalities, and combines multi-model predictions to identify the tables and columns essential for answering the natural language question. Subsequently, it outputs three distinct filtering granularities (unfiltered, table-only filtering, and full filtering) to significantly enhance the diversity and robustness of the schema representations.

(2) Few-shot Retrieval Stage:
Leveraging the schema topology information predicted in the current stage, this mechanism dynamically retrieves a fixed number of contextually relevant few-shot examples (e.g., 3 examples) from the training set that exhibit the highest semantic affinity and structural alignment, serving as in-context prompts to precisely guide the subsequent SQL deduction and generation.

(3) Candidate Generation Stage:
This stage concurrently feeds the heterogeneous schema representations, varied filtering results, raw user questions, and retrieved few-shot examples into the generator LLM. By triggering parallel greedy decoding, it efficiently constructs a diverse ensemble of candidate SQL queries, thereby successfully eliminating the reliance on expensive Chain-of-Thought (CoT) or temperature sampling.

(4) Candidate Selection Stage:
This strategy employs a confidence-aware two-stage selection mechanism. First, it applies regular majority voting across the candidate SQL queries, treating the vote distribution as a confidence metric—whereby absolute dominant results with high confidence are directly outputted. For ambiguous scenarios characterized by tied or close votes indicating low confidence, it further invokes an LLM-based pairwise ensemble voting process for the final arbitration.

\section{\ourmemory Details}
\subsection{Schema Variants Setting}
\label{app:schema}
To enrich the expression forms of database schemas and broaden the search breadth of the reasoning tree, we incorporate a diverse combination of schema formats and filtering granularities in our experiments. Specifically, we configure five core schema variant settings, as detailed in Table \ref{tab:schema}. These variants encompass MAC-Schema, M-Schema, and traditional DDL formats, each paired with one of four distinct reduction strategies: No Filtering, Column-level Filtering (Col. Filtering), Table-level Filtering (Table Filtering), and Full Filtering. By providing such heterogeneous representational designs, this setup effectively stimulates highly diverse reasoning paths, thereby fundamentally expanding the search space of the reasoning tree.

\subsection{Reasoning Tree Construction Details}
\label{app:reasoning_tree_construction}

Upon obtaining multiple diverse reasoning paths via sampling, we employ the Mermaid language as an intermediate representation to construct the structured Reasoning Tree. Mermaid is a lightweight, text-based declarative graphing language characterized by its simplistic, intuitive syntax and low parsing overhead. Its naturally text-driven format makes it exceptionally LLM-friendly, allowing the large language model to precisely articulate complex hierarchical topologies with minimal token expenditure.

Specifically, to guide the LLM in accurately mapping divergent inference trajectories into a cohesive tree structure, we manually curate a high-quality, comprehensive exemplar. This exemplar demonstrates the exact transformation from a set of candidate SQL paths containing strategic variations into standard Mermaid code, serving as a strict $\text{1-shot}$ demonstration prompt. Combined with this task-specific anchoring, the system dynamically synthesizes the target database schema, the user query, and the sampled reasoning paths into a unified instruction. The LLM then adaptively generates the corresponding Mermaid syntax, seamlessly realizing the final conflict-aware Reasoning Tree.

\begin{table}[t]
\centering
\begin{tabular}{cc}
\toprule
\textbf{Format} & \textbf{Filtering level} \\ 
\midrule
MAC Schema & No Filtering \\ 
MAC Schema & Col. Filtering \\ 
M-Schema   & Table Filtering \\ 
M-Schema   & Full Filtering \\
DDL        & Full Filtering \\ 
\bottomrule
\end{tabular}
\caption{Schema representation formats and corresponding filtering levels used in our experiments.}
\label{tab:schema}
\end{table}
\begin{table}[t]
\centering
\begin{tabular}{cc}
\toprule
\textbf{Node} & \textbf{Reasoning Step} \\ 
\midrule
$N_1$ & Question Rephrasing \\ 
$N_2$ & Table Selection \\ 
$N_3$ & Column Selection \\ 
$N_4$ & Column Function Identification \\ 
$N_5$ & Column Value Identification \\ 
$N_6$ & SQL Generation \\ 
\bottomrule
\end{tabular}
\caption{Reasoning Nodes Table in Structure-Difference Aware Reasoning Tree}
\label{tab:Nodes}
\end{table}
\begin{table*}[t]
\centering
\small
\setlength{\tabcolsep}{5pt}
\renewcommand{\arraystretch}{1.2}

\begin{tabular}{l l c c c c c c c}
\toprule

\multirow{2}{*}{\textbf{Dataset}} &
\multirow{2}{*}{\textbf{Method}} &
\multicolumn{3}{c}{\textbf{Qwen3}} &
\multicolumn{3}{c}{\textbf{Qwen2.5-Coder}} &
\textbf{Qwen3-Coder} \\

\cmidrule(lr){3-5}
\cmidrule(lr){6-8}
\cmidrule(lr){9-9}

&
&
\textbf{8B} &
\textbf{14B} &
\textbf{32B} &
\textbf{7B} &
\textbf{14B} &
\textbf{32B} &
\textbf{30B-A3B} \\

\midrule

\multirow{2}{*}{Archer-Dev}
& Baseline
& $9.62$ & $20.19$ & $25.00$
& $13.46$ & $17.31$ & $14.42$
& $32.69$ \\

& \ourmethod
& $\mathbf{12.50}$ & $\mathbf{22.12}$ & $\mathbf{27.88}$
& $\mathbf{14.42}$ & $\mathbf{23.08}$ & $\mathbf{16.35}$
& $\mathbf{36.54}$ \\

\bottomrule
\end{tabular}

\caption{Performance comparison between baseline and \ourmethod across different model scales on Archer. \textbf{Bold} indicates the best performance.}
\label{tab:archer}

\end{table*}
\begin{table}[t]
\small
\centering
\setlength{\tabcolsep}{8pt}
\renewcommand{\arraystretch}{1.2}

\begin{tabular}{l c c c}
\toprule
\textbf{Database} & \textbf{Baseline} & \textbf{\ourmethod} & $\mathbf{\Delta}$ \\
\midrule

debit\_card & $54.69$ & $\mathbf{70.31}$ & $\mathbf{+15.63}$ \\
toxicology & $58.62$ & $\mathbf{68.97}$ & $\mathbf{+10.34}$ \\
superhero  & $81.40$ & $\mathbf{91.47}$ & $\mathbf{+10.07}$ \\

\bottomrule
\end{tabular}

\caption{
Case analysis on representative databases from the BIRD dataset. The "debit\_card" database in Bird is "debit\_card\_specializing". \textbf{Bold} indicates the best performance.
}
\label{tab:database_analysis}

\end{table}
\begin{table*}[t]
\centering
\small
\begin{tabular}{l p{0.76\linewidth}}
\toprule
\multicolumn{2}{l}{\textbf{Case Study 1: SDAM Construction}} \\
\midrule

\textbf{Question ID} & 1470 \\

\textbf{DB ID} & \texttt{debit\_card\_specializing} \\

\textbf{Question} & How many gas stations in CZE has Premium gas? \\

\textbf{Reasoning Tree} &
\begin{minipage}[t]{\linewidth}
\begin{verbatim}
graph TD;
  Root[Query: Count of Gas Stations in CZE with Premium Segment]
      --> FilterCountry[WHERE Country = 'CZE']

  FilterCountry --> CheckSegment{Segment Filter}

  CheckSegment -->|Direct Match| DirectPath[Segment = 'Premium']
  DirectPath --> FinalCount[COUNT(*)]
  CandidateGood[Candidate 1,2,4,5] -.-> FinalCount

  CheckSegment -->|Join Required| JoinPath[JOIN transactions_1k & products]
  JoinPath --> BranchC{Description = 'Premium'}
  BranchC -->|Incorrect Mapping| FailPath[No Match Found (COUNT = 0)]
  CandidateBad[Candidate 3] -.-> FailPath
\end{verbatim}
\vspace{0.1mm}
\end{minipage}
\\
\textbf{Contradiction Analysis} &
The divergence occurs at the \texttt{Segment Filter} node. The correct path directly filters \texttt{Segment = 'Premium'} in the \texttt{gasstations} table, while the incorrect path introduces unnecessary joins and incorrectly treats \texttt{Premium} as a product description, leading to no matching results.
\\

\textbf{Schema Anchor} & ["table:gasstations", "column:Country", "column:Segment"]
\\

\textbf{Extracted Memory} &
To count gas stations in CZE with Premium segment, filter the \texttt{gasstations} table using \texttt{Country = 'CZE'} and \texttt{Segment = 'Premium'}.
\\

\bottomrule
\end{tabular}
\caption{A case study of the \ourmemory construction process, illustrating how the model performs Structure-Difference Aware Reasoning Tree analysis, Contradiction-Aware Reflection, and Schema-Grounded Memory Evolution to construct SDAM.}
\label{tab:case_study1}
\end{table*}
\begin{table*}[t]
\centering
\small
\begin{tabular}{p{0.17\linewidth} p{0.80\linewidth}}
\toprule
\multicolumn{2}{l}{\textbf{Case Study 2: SDAM Usage}} \\
\midrule

\textbf{Question ID} & 1474 \\

\textbf{DB ID} & \texttt{debit\_card\_specializing} \\

\textbf{Question} & Which customers, paying in CZK, consumed the most gas in 2011? \\

\textbf{Incorrect SQL (w/o Memory)} &
\begin{minipage}[t]{\linewidth}
\begin{verbatim}
SELECT *
FROM (
    SELECT "customer_id"
    FROM (
        SELECT CASE
            WHEN SUM("amount") > 0
            THEN 'positive'
            ELSE 'negative'
        END AS status
    )
) AS t
WHERE 1 = 1
\end{verbatim}
\vspace{0.1mm}
\end{minipage}
\\

\textbf{Retrieved Memory} &
To identify customers who consumed the most gas in 2011, aggregate consumption per customer using \texttt{GROUP BY CustomerID} and \texttt{SUM(Consumption)} from the \texttt{yearmonth} table, filtering for dates in 2011. 
Additionally, join the \texttt{yearmonth} table with the \texttt{customers} table on \texttt{CustomerID} and filter \texttt{Currency = 'CZK'}.
\\

\textbf{Correct SQL} &
\begin{minipage}[t]{\linewidth}
\begin{verbatim}
SELECT T1.CustomerID
FROM yearmonth AS T1
INNER JOIN customers AS T2
ON T1.CustomerID = T2.CustomerID
WHERE T2.Currency = 'CZK'
AND T1.Date BETWEEN 201101 AND 201112
GROUP BY T1.CustomerID
ORDER BY SUM(T1.Consumption) DESC
LIMIT 1
\end{verbatim}
\vspace{0.1mm}
\end{minipage}
\\

\bottomrule
\end{tabular}
\caption{A case study of SDAM usage, showing how retrieved SDAM guides the model from an incorrect SQL structure to the correct SQL query.}
\label{tab:case_study2}
\end{table*}
\subsection{Detailed Mechanism of Contradiction Detection}
\label{app:con_det}
To enable the proposed \ourmemory to effectively capture and evolve structured semantic knowledge from instance-level errors, we design a comprehensive contradiction detection pipeline. During the reasoning and evolution stages, contradictions are systematically identified across three primary dimensions:
\begin{itemize}[leftmargin=*,nosep]
    \item \textbf{Inter-Structure Contradictions within the Reasoning Tree:} This dimension focuses on detecting logical divergences between alternative reasoning paths. By exploring two or more sibling branches originating from the same parent node within the reasoning tree, the system identifies conflicts in execution strategies. For instance, one branch might employ a float conversion before a division operation (e.g., \texttt{CAST(total\_score AS FLOAT) / count}), whereas another sibling branch directly performs integer division. Identifying such calculative discrepancies allows the framework to explicitly weigh the structural validity of different reasoning paths.
    \item \textbf{Semantic-Schema Mismatches:} These contradictions occur when the generated candidate SQL query violates the underlying constraints or data formats defined in the database schema. This encompasses column-level inconsistencies caused by shallow surface semantics (e.g., referencing \texttt{Student\_Name} while the schema explicitly defines it as \texttt{Full\_Name}, which triggers execution failures), as well as data-format discrepancies where the SQL filters rows using a date string (e.g., \texttt{'2024-01-01'}) while the database stores chronological data as an integer representation (e.g., \texttt{20240101}), yielding incorrect empty results.
    \item \textbf{Execution-Intent Contradictions:} Even when a candidate SQL query is syntactically valid and executes successfully, its final return values may conflict with the user's natural language intent. This manifests either as a value mismatch that fails to address the quantitative target of the question (e.g., returning the ``total sum'' when the user inquires about the ``average score''), or as a column omission that fails to return the full set of requested attributes (e.g., projecting only school names via \texttt{SELECT name} when the query explicitly demands to ``list the name and address of all schools'').
\end{itemize}

\section{Supplementary Experiment}
\subsection{Experiments on Archer Dataset}
\label{app:exp_arc}
To further evaluate the generalization performance of \ourmethod in highly sophisticated reasoning scenarios, we conduct extended experiments on the \textbf{Archer} \cite{zheng2024archer} benchmark. Archer is a cross-domain bilingual dataset specifically designed for complex reasoning, comprising 1,042 English and Chinese questions respectively, paired with 521 unique SQL queries. By explicitly incorporating arithmetic, common-sense, and hypothetical reasoning, Archer introduces a substantial shift in structural and logical complexity. Experimental results (Table \ref{tab:archer}) reveal that \ourmethod delivers its most pronounced performance leap on this challenging benchmark. Notably, when deployed on the Qwen2.5-Coder-14B backbone, our method drives a substantial execution accuracy (EX) improvement of \textbf{5.77}\%, while on the strongest backbone, the performance scales up to a peak of \textbf{36.54\%}. This substantiates that the historical expertise accumulated via our structured memory evolution effectively empowers LLMs to navigate severe multi-step cross-domain reasoning bottlenecks.

\subsection{Single-Database Analysis}
\label{app:exp_sin}

Table \ref{tab:database_analysis} presents the evaluation results of \ourmethod across several representative individual databases. Incorporating the memory evolution framework leads to significant performance boosts: the EX accuracy on \texttt{debit\_card\_specializing} escalates from $54.69\%$ to $70.31\%$ (a \textbf{15.63}\% gain), while \texttt{toxicology} and \texttt{superhero} exhibit improvements of \textbf{10.34}\% and \textbf{10.07}\%, respectively.

Analysis indicates that these databases generally lack comprehensive schema documentation and feature highly specialized domain-specific semantics (e.g., credit card transactions or toxicology). Consequently, conventional methods are highly susceptible to semantic errors by on surface-level schema information. In contrast, leveraging contradiction-driven memory distillation, \ourmethod progressively abstracts robust domain rules and implicit constraints from erroneous cases. Reusing such deep semantic knowledge effectively guides downstream complex reasoning and schema item selection, significantly boosting the model's generalization capabilities in specialized professional scenarios.

\section{Case Study}
\label{app:case_study}
\subsection{SDAM Construction}

Table \ref{tab:case_study1} presents a case study of the SDAM construction process. The model first builds a Structure-Difference Aware Reasoning Tree by generating multiple candidate reasoning paths. The reasoning divergence occurs at the \texttt{Segment Filter} node, where the incorrect branch introduces unnecessary joins and mistakenly treats \texttt{Premium} as a product description. 

Based on this divergence, Contradiction-Aware Reflection identifies the semantic inconsistency between \texttt{products.Description} and gas station segment information. Finally, the Schema-Grounded Memory Evolution mechanism binds the extracted knowledge to schema anchors such as \texttt{gasstations.Country} and \texttt{gasstations.Segment}, producing reusable schema-grounded memory for SQL generation.

\subsection{SDAM Usage}

Table \ref{tab:case_study2} presents a case study of SDAM usage during SQL generation. Without memory, the baseline model produces an invalid SQL query that fails to capture the schema structure and task semantics. Specifically, the generated SQL contains irrelevant aggregation logic and does not correctly model the relationship between customers, currency constraints, and yearly gas consumption.

After retrieving relevant SDAM, the model successfully recalls reusable reasoning patterns, including aggregating \texttt{SUM(Consumption)} grouped by \texttt{CustomerID}, filtering records in 2011, and joining the \texttt{customers} and \texttt{yearmonth} tables through \texttt{CustomerID}. Guided by the retrieved schema-grounded memory, the model generates the correct SQL query and accurately identifies the target customer. This case demonstrates that SDAM can effectively transfer historical reasoning experience to subsequent Text-to-SQL tasks, thereby improving structural reasoning and schema alignment.

\section{Prompts}
\label{app:prompts}
In this section, we present the explicit prompt templates utilized during \ourmemory construction. To ensure rigorous reproducibility, all system prompts are detailed in their original configurations. 

Specifically, the instruction for Candidate Generation, Contradiction-Aware Reflection, Memory Extraction, and Schema-Grounded Memory Filtering are illustrated in Figure \ref{pro:gen}, Figure \ref{pro:ref}, Figure \ref{pro:ext}, and Figure \ref{pro:fil}, respectively.
\begin{figure*}[t]
    \centering
    \small
    \input{prompts/generator}
    \caption{Generator Prompt}
    \label{pro:gen}
\end{figure*}

\begin{figure*}[t]
    \centering
    \small
    \input{prompts/reflector}
    \caption{Reflector Prompt (Simplified)}
    \label{pro:ref}
\end{figure*}

\begin{figure*}[t]
    \centering
    \small
    \input{prompts/extractor}
    \caption{Extractor Prompt}
    \label{pro:ext}
\end{figure*}

\begin{figure*}[t]
    \centering
    \small
    \input{prompts/filter}
    \caption{Filter Prompt}
    \label{pro:fil}
\end{figure*}

\end{document}